\documentclass[letterpaper, 10 pt, conference]{ieeeconf}

\IEEEoverridecommandlockouts                              %

\usepackage[table]{xcolor}
\usepackage{colortbl}

\usepackage{times}
\usepackage{multicol}

\usepackage{color,soul}
\usepackage{graphicx}
\usepackage{wrapfig}
\usepackage{siunitx}
\usepackage{bbold}
\usepackage{subcaption}
\usepackage{dsfont}
\usepackage{xspace}
\usepackage[bookmarks=true]{hyperref}

\usepackage{adjustbox}
\usepackage{varwidth}

\usepackage{array} %
\usepackage{tabularx}
\usepackage{multirow}
\usepackage{multicol}
\usepackage{booktabs}

\usepackage{algorithm}
\usepackage{algpseudocode}
\usepackage{amssymb}
\usepackage{amsmath}
\usepackage{cite}
\usepackage{wrapfig}
\usepackage{pifont}
\usepackage{colortbl}

\newcommand{\modelname}{\textsc{Doduo}}

\newlength\savewidth\newcommand\shline{\noalign{\global\savewidth\arrayrulewidth
  \global\arrayrulewidth 1pt}\hline\noalign{\global\arrayrulewidth\savewidth}}
\newcommand{\tablestyle}[2]{\setlength{\tabcolsep}{#1}\renewcommand{\arraystretch}{#2}\centering\footnotesize}

\definecolor{tablegray}{gray}{0.7}
\definecolor{cerulean}{rgb}{0.0, 0.48, 0.65}

\title{\LARGE \bf \modelname: Learning Dense Visual Correspondence from \\ Unsupervised Semantic-Aware Flow}

\author{
    Zhenyu Jiang$^{1}$, Hanwen Jiang$^{1}$, Yuke Zhu$^{1}$
    \thanks{
        $^{1}$ Department of Computer Science, the University of Texas at Austin. Correspondance to {\tt\small zhenyu@utexas.edu}
    }   
}

\begin{document}
\maketitle

\begin{abstract}
Dense visual correspondence plays a vital role in robotic perception. This work focuses on establishing the dense correspondence between a pair of images that captures dynamic scenes undergoing substantial transformations. We introduce \modelname{} to learn general dense visual correspondence from in-the-wild images and videos without ground truth supervision. Given a pair of images, it estimates the dense flow field encoding the displacement of each pixel in one image to its corresponding pixel in the other image. \modelname{} uses flow-based warping to acquire supervisory signals for the training. Incorporating semantic priors with self-supervised flow training, \modelname{} produces accurate dense correspondence robust to the dynamic changes of the scenes. Trained on an in-the-wild video dataset, \modelname{} illustrates superior performance on point-level correspondence estimation over existing self-supervised correspondence learning baselines. We also apply \modelname{} to articulation estimation and zero-shot goal-conditioned manipulation, underlining its practical applications in robotics. Code and additional visualizations are available at \href{https://ut-austin-rpl.github.io/Doduo/}{\textcolor{cerulean}{\url{https://ut-austin-rpl.github.io/Doduo/}}}
\end{abstract}

\section{Introduction}

\begin{figure}[t]
    \centering
    \includegraphics[width=\linewidth]{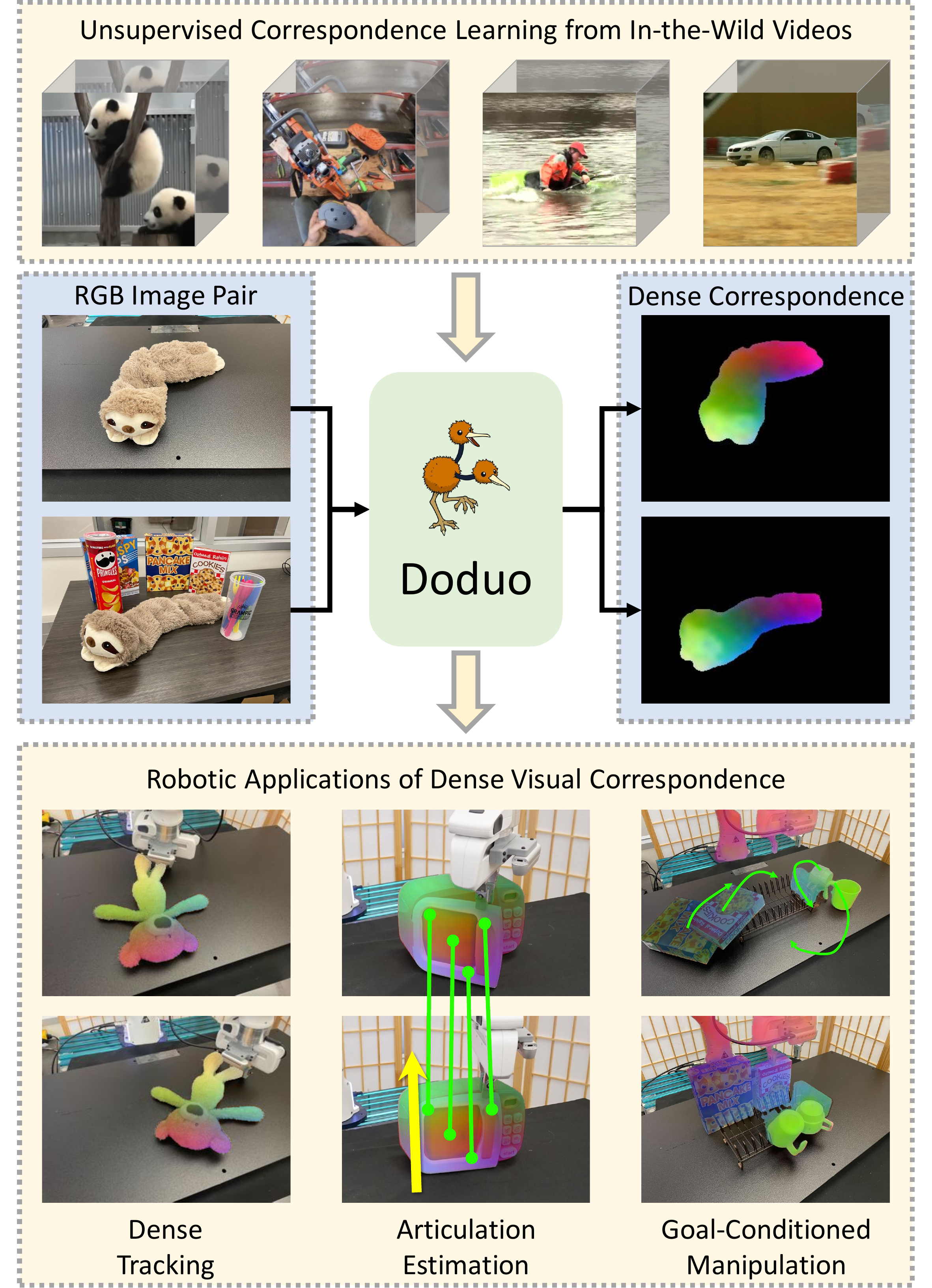}
    \caption{\small{\modelname{} is trained on in-the-wild videos without manual annotations. It can predict accurate dense correspondence from pairs of images to enable a diverse set of robotic applications.}}
    \vspace{-0.2in}
    \label{fig:over-view}
\end{figure}

Dense visual correspondence involves identifying the corresponding pixel in a target image for any given pixel in a source image. Correspondence is a cornerstone of robot perception, driving a multitude of robotic applications, such as dense visual tracking~\cite{newcombe2011dtam,izadi2011kinectfusion,newcombe2011kinectfusion,whelan2015elasticfusion,henry2012rgb}, articulation estimation~\cite{Black1996EigenTrackingRM, Hausman2015ActiveAM, jiang2022ditto}, and deformable object modeling~\cite{Yin2021ModelingLP, Tulsiani2017LearningCD, Liao2009ModelingDO}. These applications typically require fine-grained correspondence reasoning of \emph{dynamic scenes undergoing substantial transformations}.

In recent years, supervised deep learning methods~\cite{teed2020raft, LoweDavid2004DistinctiveIF, Sarlin2019SuperGlueLF, Yi2016LIFTLI, Dusmanu2019D2NetAT, Choy2016UniversalCN, Sun2021LoFTRDL, Jiang2021COTRCT} have made great strides in learning correspondences from annotated datasets. These methods perform well in constrained scenarios, such as with rigid objects/scenes and minor viewpoint changes. However, their efficacy under large, non-rigid motions is curtailed by the prohibitive costs (thus limited availability) of ground-truth annotations. 
Alternatively, research exemplified by DenseObjectNets~\cite{schmidt2016self,florence2018dense,florence2019self,manuelli2020keypoints,yen2022nerf} employs robot interaction to capture multi-view observations of objects and generate correspondence ground truth for training. However, such procedures only yield category-level visual descriptors that fail to generalize beyond the category of the captured objects.
Self-supervised learning methods like DINO~\cite{Caron2021EmergingPI} enable large model training on in-the-wild images and videos, offering a more scalable and generalizable approach to semantic correspondence learning without manual annotations. Nevertheless, how to learn a generalizable model to robustly establish fine-grained correspondence required by robotic tasks remains an open question.

Two notable categories of self-supervised techniques have been studied in the visual correspondence learning literature. The first category is \textbf{self-supervised optical flow}~\cite{Long2016LearningIM,meister2018unflow,ren2017unsupervised,yu2016back,liu2019selflow,huang2023self}. The key idea revolves around predicting dense optical flow between adjacent video frames and utilizing a photometric loss to minimize the color discrepancy between the matched pixels. The pixel-level supervision from the photometric loss allows for learning fine-grained correspondences. However, the color-matching property of the photometric loss limits the model's resilience against large appearance variations and changes in illumination. The second category is \textbf{self-supervised semantic feature learning}~\cite{Caron2021EmergingPI,Oquab2023DINOv2LR}, which generates semantically consistent feature representations and exhibits robustness to large lighting and viewpoint variations. Despite this, in cases where pixels have similar semantics, such as points on a cabinet door, the estimated correspondence is coarse due to indistinguishable semantics. These two categories of methods possess complementary strengths and weaknesses. The former offers satisfying fine-grained correspondence but struggles to generalize under appearance changes, while the latter provides robust coarse matching. We argue that a more general correspondence learning method should integrate the advantages of both categories.

To this end, we introduce \modelname{} (\underline{\textbf{D}}ense Visual C\underline{\textbf{o}}rrespon\underline{\textbf{d}}ence from \underline{\textbf{U}}nsupervised Semantic-Aware Fl\underline{\textbf{o}}w), designed to efficiently learn fine-grained dense correspondences between a pair of images with substantial variations.
Using a transformer-based neural network~\cite{vaswani2017attention,liu2021swin}, \modelname{} extracts correspondence-aware feature maps from the image pairs and estimates the flow from source to target based on feature similarities.
Learning to find correspondence from scratch with all the pixels in the target image as matching candidates is difficult, especially when direct supervision of ground truth correspondence is not available. We mitigate this challenge by incorporating robust semantic features from DINO~\cite{Caron2021EmergingPI}.
When finding the matching of a query point in the source image, we use DINO feature correspondence to select a subset of points on the target image with higher semantic similarities to the query point. Incorporating the coarse correspondence from DINO allows our model to identify fine-grained correspondences inside this subset of matching candidates.

One noteworthy challenge in self-supervised flow-based methods is the occlusion across frames. We tackle this issue by using predicted flow to select region masks from an off-the-shelf image segmentation model, thereby identifying common regions visible in both images. As our correspondence model improves, the visible region localization becomes more precise and, in turn, facilitates model training.

We train \modelname{} on frames from an in-the-wild Youtube-VOS video dataset~\cite{Xu2018YouTubeVOSAL} and evaluate it on the TAP-Vid~\cite{Doersch2022TAPVid}, a benchmark with fine-grained point correspondence annotations. Our model demonstrates superior performance across all metrics compared with baselines. Furthermore, we apply \modelname{} to the articulation estimation task and show its ability to estimate articulation without the need for training with any ground truth articulated data. We further conduct real-robot experiments of zero-shot goal-conditioned manipulation, highlighting how accurate dense visual correspondence enables precise actions in fine-grained object manipulation. These results affirm the broad applicability of \modelname{} to downstream robotic tasks without the need for training with domain-specific data.

\section{Related Work}

\begin{figure*}[t]
    \centering
    \includegraphics[width=0.75\linewidth]{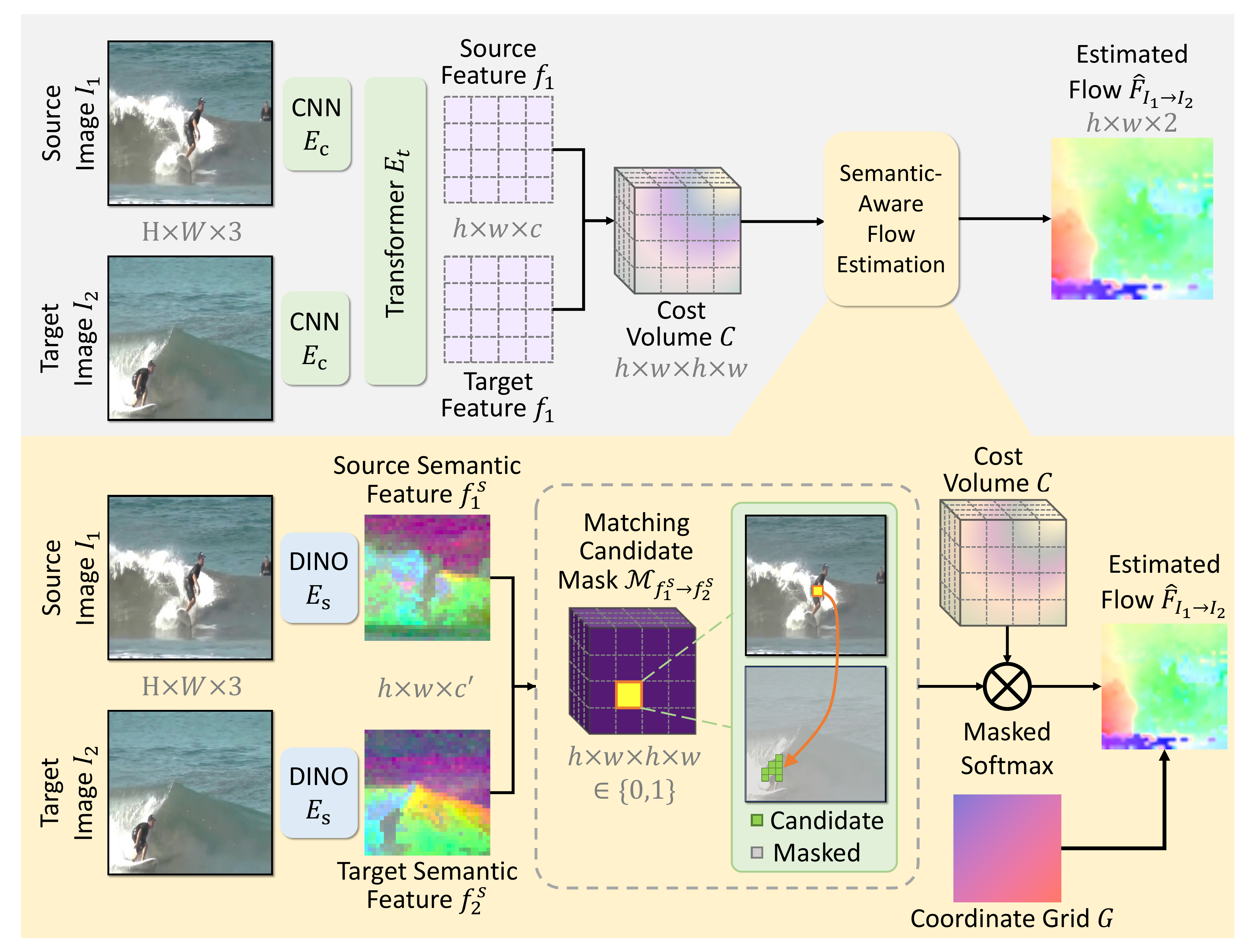}
    \caption{\small{\textbf{\modelname{} Model Architecture.} Our model takes two images as inputs and uses a Transformer-based network to extract features of both images. Then, we construct a 4D cost volume of all pairs of feature pixels and predict a dense semantic-aware flow field given the cost volume. For semantic-aware flow estimation, we use a DINO encoder~\cite{Caron2021EmergingPI} to extract semantic feature maps of both input frames. According to the similarity of the semantic feature map, we compute a matching candidate mask for each of the feature points of the source frame and integrate this information during flow estimation using masked Softmax.}}
    \vspace{-0.2in}
    \label{fig:pipeline}
\end{figure*}

\noindent\textbf{Unsupervised Visual Correspondence Learning.} Recent years have seen considerable advancements in unsupervised visual correspondence learning, which capitalizes on a plethora of unannotated data for training~\cite{Zhou2017UnsupervisedLO, Wang2020LearningFD, Schmidt2017SelfSupervisedVD, Vondrick2018TrackingEB, wang2019learning, Wang2019UnsupervisedDT, Zhong2017SelfSupervisedLF, ElBanani2022SelfsupervisedCE}. Despite the progress, the practical applicability of these models remains limited due to various inherent model design constraints.
Most unsupervised visual correspondence learning frameworks belong to three main types. 

The first type of method learns from synthesized self-supervision~\cite{Truong2021WarpCF, Truong2022ProbabilisticWC}. These methods generate synthetic warping between image pairs and harness the reverse warping as the supervisory signal. However, its inability to simulate the dynamic aspects of real-world scenes, such as moving objects, results in suboptimal performance.

The second is unsupervised optical flow~\cite{liu2019selflow, wang2018occlusion,wang2019learning,Wang2020LearningFD,Janai2018UnsupervisedLO,yu2016back,Jonschkowski2020WhatMI, stone2021smurf}. These methods can deal with dynamic scenes, as they learn from natural motions in the videos. However, they train models with the photometric loss, assuming the matched pixels possess the same color. This assumption requires the training image pairs to have small illuminance changes, typically neighboring frames in a video. Consequently, they struggle to handle image pairs with significant appearance or content changes.

The third is the semantic representation learning~\cite{Xu2021RethinkingSC, Caron2021EmergingPI, Oquab2023DINOv2LR, Xiong2021SelfSupervisedRL,Jabri2020SpaceTimeCA, Vondrick2018TrackingEB, wang2019learning, Lai2020MASTAM,Lai2019SelfsupervisedLF, Hu2022SemanticAwareFC}. To learn semantic features, these methods leverage self-supervised learning objectives, including contrastive learning, self-distillation, and cycle consistency. These features are robust to appearance variations, including illumination, viewpoint, and spatial transformations. Nevertheless, features of points that belong to the same semantic regions tend to be indistinguishable, resulting in coarse correspondences in these regions. Such coarse matching is insufficient for downstream robotics applications.

\vspace{1mm}
\noindent\textbf{Visual Descriptor Learning from Interaction.}
There is a branch of research using robot interaction to learn dense visual descriptors~\cite{schmidt2016self,florence2018dense,florence2019self,manuelli2020keypoints,yen2022nerf,graf2023learning}. DenseObjectNets~\cite{florence2018dense} is one of the pioneer works in this direction. DenseObjectNets uses a camera mounted on a robot arm to collect multi-view observations and extract ground truth correspondence with 3D reconstruction. The learned visual descriptors are robust to variations in object pose, lighting, and deformation. However, the learned descriptor cannot generalize to objects from novel categories, limiting its applicability to general robot manipulation tasks with diverse object categories.

In contrast to existing works, \modelname{} learns a general correspondence model that predicts dense correspondence between image pairs capturing scenes undergoing substantial variations. Since it is trained on large-scale in-the-wild frame pairs containing diverse visual contents, \modelname{} generalizes to different objects and scenes without finetuning.

\section{Method}

We now present \modelname{}, a self-supervised learning approach for dense visual correspondence based on semantic-aware flow. Fig.~\ref{fig:pipeline} illustrates the \modelname{} model. We introduce a semantic-aware flow estimation model (Sec.~\ref{sec: flow prediction}), leveraging robust coarse matching from pre-trained semantic features for correspondence prediction. Moreover, to avoid the ill-posed problem of estimating the flow of occluded pixels, we introduce a bootstrapping strategy to locate visible regions in both images and apply supervision only on the visible regions (Sec.~\ref{sec: visible_region_discovery}).
In addition, we introduce a novel pixel-level feature-metric loss as the self-supervision for training \modelname{} (Sec.~\ref{sec: loss}). The pixel-level losses encourage the model to predict fine-grained correspondences. 

\subsection{Flow Estimation}
\label{sec: flow prediction}

\noindent\textbf{Problem Formulation.}
We study the problem of pixel-level dense correspondence between a source image $I_1 \in \mathbb{R}^{H \times W \times 3}$ and a target image $I_2 \in \mathbb{R}^{H \times W \times 3}$. Given the pair of input images $I_1, I_2$, we aim to predict a flow field $F_{I_1 \rightarrow I_2}$.
The flow field $F_{I_1 \rightarrow I_2}$ contains a 2D vector for each pixel in the source image $I_1$, representing the offset from the pixel coordinate in $I_1$ to the corresponding pixel coordinate in $I_2$, as $F_{I_1 \rightarrow I_2}(p^1_x, p^1_y) = (p^2_x - p^1_x, p^2_y - p^1_y)$.
$(p^1_x, p^1_y)$ stand for the pixel coordinates of a point $p^1$ in image $I_1$, while $(p^2_x, p^2_y)$ denote the pixel coordinates of the corresponding point $p^2$ in image $I_2$.

\vspace{1mm}
\noindent\textbf{Cost Volume Prediction.}
As shown in Fig.~\ref{fig:pipeline} (top), \modelname{} first extracts image features from $I_1, I_2$ with a convolutional encoder independently. We then use a Transformer-based neural network to correlate their image features. The output image features of the two images $I_1$ and $I_2$ are denoted as $f_1$ and $f_2$, respectively. $f_1$, $f_2 \in \mathbb{R}^{h\times w\times c}$, $h = H/8$ and $w = W/8$.
Subsequently, we compute the cost volume between two frames based on feature similarity $C = \frac{f_1 f_2^T}{\sqrt{c}} \in \mathbb{R}^{h \times w \times h \times w}$.
The cost volume $C$ represents the similarity of each pixel pair between the source and target images.

\begin{figure*}
    \centering
    \includegraphics[width=0.85\linewidth]{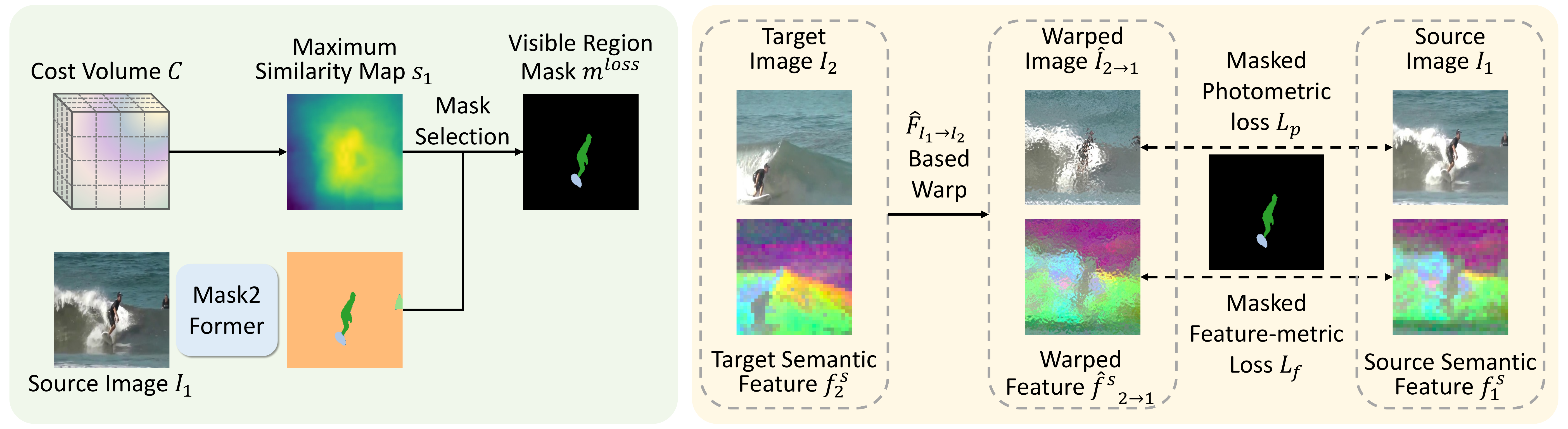}
    \caption{\small{\textbf{\modelname{} Training.} (Left) Bootstrapping Visible Region Discovery. We use Mask2Former~\cite{Cheng2021Mask2FormerFV} to get object segments of the source frame and select the segments that are most likely to be matched in the target frame using the predicted cost volume. (Right) We use the estimated flow to warp the pixels and the DINO features of the target image. The training objective is to minimize the photometric and feature-metric loss between the warped image and the source image in located visible regions.}}
    \vspace{-0.2in}
    \label{fig:training}
\end{figure*}

\vspace{1mm}
\noindent\textbf{Semantic-Aware Flow Estimation.}
\label{sec:flow}
We exploit the coarse semantic correspondence from the DINO image encoder~\cite{Caron2021EmergingPI} to provide priors for learning fine-grained correspondence. 
As shown in Fig.~\ref{fig:pipeline} (bottom), we use DINO to construct a matching candidate mask.
The matching candidate mask is applied to the cost volume, narrowing down the matching space for dense flow estimation.
In detail, we use the DINO encoder ${E}_s$ to acquire semantic feature maps
$f^{s}_{i} = {E}_s (I_i), i=1,2$ and $f^{s}_{i}\in \mathbb{R}^{h \times w \times c'}$.
For each feature pixel $p \in f^{s}_{1}$, we identify the top $1\%$ similar pixels in $f^{s}_{2}$. The resulting feature pixels in $f^{s}_{2}$ constitute a mask for the feature pixel $p$, denoted as $\mathcal{M}_p\in \{0, 1\}^{h \times w}$, representing the matching candidates of $p$ in $f^{s}_{2}$. We compute the mask for each $p \in f^{s}_1$ to obtain a matching candidate mask $\mathcal{M}_{f^{s}_{1} \rightarrow f^{s}_{2}} \in \{0, 1\}^{h \times w \times h \times w}$.

Next, we apply a masked softmax to the last two dimensions (belonging to the target image $I_2$) of the cost volume $C$ with the matching mask $\mathcal{M}_{f^{s}_{1} \rightarrow f^{s}_{2}}$, yielding a normalized matching distribution $\tilde{C} = \textrm{MaskedSoftmax}(C$$, \mathcal{M}_{f^{s}_{1} \rightarrow f^{s}_{2}})$, where $\tilde{C} \in \mathbb{R}^{h\times w}$. The dense correspondence $G'$ can then be attained by calculating the weighted average of the matching distribution with the 2D coordinates of pixel grid $G \in \mathbb{R}^{h \times w \times 2}$. And the flow field can be derived as the difference between the corresponding pixel coordinates as $\hat{F}_{I_1 \rightarrow I_2} = \tilde{C}G - G \in \mathbb{R}^{h \times w \times 2}$.
Finally, we use bilinear upsampling to obtain a flow field of original resolution $\hat{F}^{up}_{I_1 \rightarrow I_2} \in \mathbb{R}^{H \times W \times 2}$.

\subsection{Bootstrapping Visible Region Discovery}
\label{sec: visible_region_discovery}
Self-supervised flow prediction contends with the challenge of occlusions. Applying loss on occluded pixels encourages the network to correspond pixels devoid of genuine matchings in the other frame, which in turn diminishes its performance\cite{Wang2017OcclusionAU, Brox2004HighAO, Janai2018UnsupervisedLO}.
Instead of detecting occlusions, we introduce a method of identifying instance-level visible regions in both frames. The visible regions are used as a mask for computing the loss, bringing instance-level priors for correspondence learning.
This technique is particularly beneficial for image pairs exhibiting significant differences where occluded pixels are prevalent. 

As shown in Fig.~\ref{fig:training} (left), we employ an off-the-shelf image segmentation model Mask2Former~\cite{Cheng2021Mask2FormerFV} to get image region masks.
Specifically, we attain segmentation masks of $I_1$ as $m_1 = \textrm{Mask2Former}(I_1) \in \{0, 1\}^{H \times W \times N}$, where $N$ is the maximum number of image regions. 

To identify regions that are most likely to be matched in $I_2$, we leverage the cost volume $C$ produced by \modelname{}. For each feature point of $f_1$, we compute the maximum similarity score, the similarity between this feature point and the closest feature point in $f_2$. A higher maximum similarity score indicates that the feature point is more likely to be matched in $f_2$. We obtain the maximum similarity map by maximizing out the last two dimensions of the cost volume $s_1 = \max(C, \text{dim}=(3,4)) \in \mathbb{R}^{H \times W}$. Next, we average the maximum similarity score for each pixel in a segment.
We identify the top $k$ segments with the largest scores, where the resulting segments are visible regions between the two images with high possibility, denoted as $m^{loss} \in \{0, 1\}^{H \times W \times k}$.

\begin{figure*}
    \centering
    \vspace{0.04in}
    \includegraphics[width=0.9\linewidth]{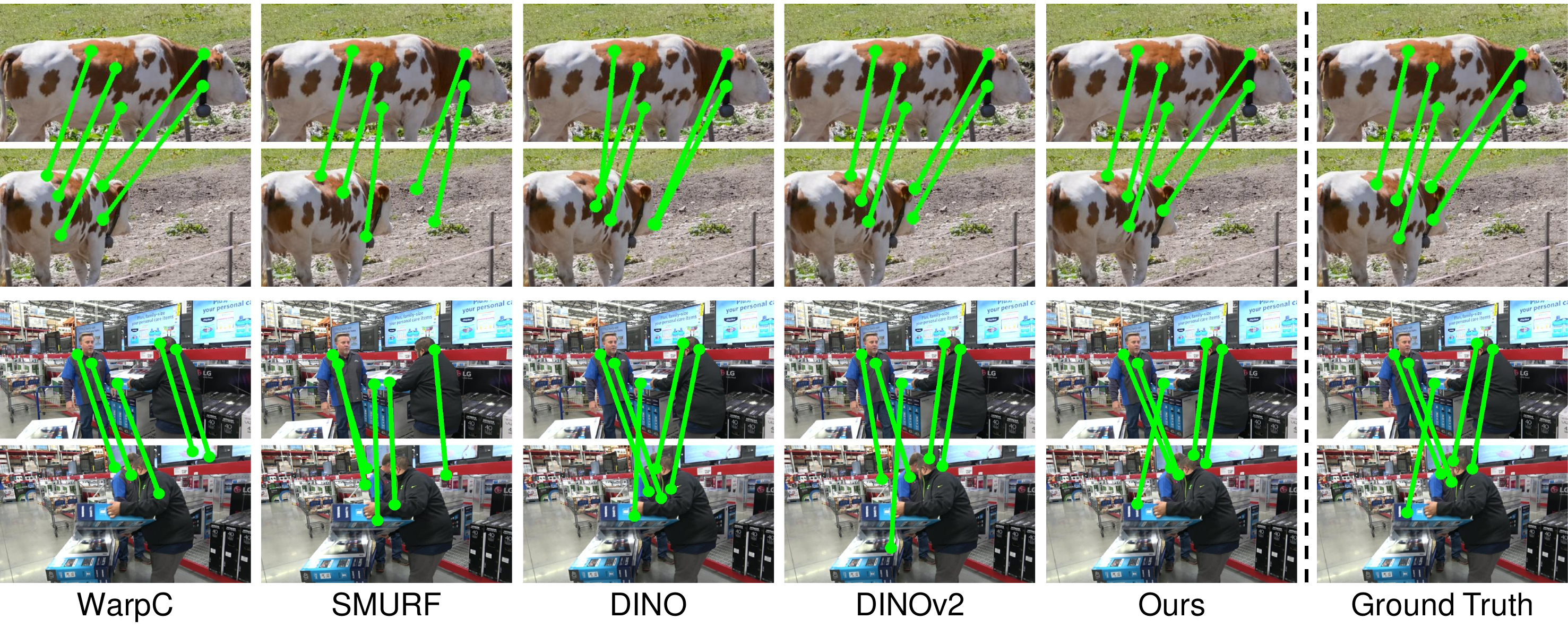}
    \vspace{-0.05in}
    \caption{\small{Visualization of pixel-level correspondence results on TAP-Vid DAVIS dataset.}}
    \vspace{-0.1in}
    \label{fig:tapvid-qual}
\end{figure*}

\subsection{Self-Supervised Losses}
\label{sec: loss}
\noindent\textbf{Photo-Metric and Feature-Metric Losses.} 
As shown in Fig.~\ref{fig:training} (right), following self-supervised optical training, we warp $I_2$ with the predicted flow field to obtain $\hat{I}_{2 \rightarrow 1} = \text{warp}(I_2, \hat{F}_{I_1 \rightarrow I_2})$. We then employ the photo-metric loss to minimize the difference between $\hat{I}_{2 \rightarrow 1}$ and $I_1$. Specifically, we only apply the loss on the visible regions in both frames, identified as mask $m^{loss}$. And we use the Charbonnier loss $\psi$~\cite{Charbonnier1994TwoDH} to calculate the pixel-level discrepancy. Photometric loss offers fine-grained pixel-level supervision. However, it presumes color consistency between frames. This assumption holds for neighboring video frames but falters for frame pairs with appearance changes. Consequently, we introduce a feature-metric loss to motivate the network to concentrate on matching semantic features, which is more robust under appearance changes. We warp the semantic feature map of $I_1$ with the predicted flow field to get $\hat{f^{s}}_{2 \rightarrow 1} = \text{warp}(f^{s}_{2}, \hat{F}_{I_1 \rightarrow I_2})$. Next, we acquire photo-metric loss $L_p$ and feature-metric loss $L_f$ by computing $\psi(I_1 - \hat{I}_{2 \rightarrow 1})$ and $\psi(f^{s}_{1} - \hat{f^{s}}_{2 \rightarrow 1})$ inside the visible regions $m^{loss}$.

\vspace{1mm}
\noindent\textbf{Flow Regularizer.}
Prior works on self-supervised optical flow~\cite{Yu2016BackTB, Jonschkowski2020WhatMI} regularize the flow prediction using a smoothness term, assuming the motion between image pairs is translational. To relax the unrealistic assumption, we propose the distance consistency loss.
We enforce the distance between neighboring pixels to remain consistent after warping by the flow. For a pair of neighboring pixels $(p^1_i, p^1_j)$ in $I_1$, we find the corresponding pixels, denoted as $(\hat{p}^2_i, \hat{p}^2_j)$, in $I_2$ using our estimated flow field.
We minimize $D(i, j) = \psi(||p^1_j - p^1_i|| - ||\hat{p}^2_j - \hat{p}^2_i||)$ inside different identified image regions and acquire regularization loss $L_d$. Our final training loss is defined as
$L = L_p + L_f + L_d$.

\vspace{1mm}
\noindent\textbf{Training \modelname{}.}
We train \modelname{} on video frames from Youtube-VOS dataset~\cite{Xu2018YouTubeVOSAL}. We randomly choose frame pairs with temporal intervals between 1-3 seconds for controllable variations between inputs. We apply random crop augmentation on the chosen frames.

\section{Experiments}

We first introduce the baselines for comparison.
We compare \modelname{} with WarpC~\cite{Truong2021WarpCF}, SMURF~\cite{stone2021smurf}, and DINO~\cite{Caron2021EmergingPI}. WarpC is an unsupervised warp-based method that predicts the dense flow field. SMURF is a method for unsupervised learning of optical flow.
DINO (and recently updated version DINOv2~\cite{Oquab2023DINOv2LR}) is a self-supervised semantic representation learning method. For DINO, we predict the dense flow in the same way as \modelname{} (Sec.~\ref{sec:flow}).

\subsection{Evaluation Datasets and Metrics}
We evaluate \modelname{} on TAP-Vid~\cite{Doersch2022TAPVid} dataset and D3D-HOI~\cite{Xu2021D3DHOID3} dataset for evaluating fine-grained visual correspondence and articulation estimation, respectively.

\vspace{1mm}
\textbf{TAP-Vid DAVIS} is a long-term point-tracking dataset containing
30 videos from the DAVIS dataset~\cite{PontTuset2017DAVIS} and providing point-level annotations. For each point, we use the frame where it first appears as the source image and use each of the following frames as the target image. 
We use the predicted flow to warp the source image points to the target image. The warped points are considered as their corresponding points in the target image. 
We follow the official metrics for evaluation, including 1) \textit{Average Distance} (AD) between the prediction and the ground truth in pixels; 2)  $<\delta^x_{avg}$, which measures the average percentage of points within the distance threshold of 1, 2, 4, 8, and 16 pixels; 3) \textit{Average Jaccard} (AJ), measuring the precision under the mentioned distance thresholds.

\begin{table}[t]
    \centering
    \tablestyle{6pt}{1.1}
    \scriptsize
    \setlength\tabcolsep{3pt}
    \begin{tabular}{l|ccc|cccc}
    \shline
    & \multicolumn{3}{c|}{Fine-Grained Corr.} & \multicolumn{4}{c}{Articulation Estimation Error} \\
    \cmidrule(lr){2-4}\cmidrule(lr){5-8}
  & AJ$_\uparrow$  & AD$_\downarrow$ & $< \delta^x_{avg\uparrow}$ & Angle$_\downarrow$ & Pos$_\downarrow$ & State$_\downarrow$ & Dist$_\downarrow$ \\
  \hline
  WarpC~\cite{Truong2021WarpCF} & 25.8 & 28.1 & 35.8 & 11.41 & 0.182 & 13.43 & 0.148 \\
  SMURF~\cite{stone2021smurf} & 29.8 & 27.4 & 42.7 & 10.66 & 0.130 & 8.94 & 0.132 \\
  DINO~\cite{Caron2021EmergingPI} & 25.7 & 15.2 & 35.5 & 10.29 & 0.161 & 11.56 & 0.146 \\
  DINOv2~\cite{Oquab2023DINOv2LR} & 27.2 & 13.4 & 36.0 & 12.14 & 0.144 & 8.06 & 0.126 \\
  \modelname{} (ours) & \textbf{33.2} & \textbf{12.3} & \textbf{43.5} & \textbf{9.60} & \textbf{0.111} & \textbf{6.10} & \textbf{0.110} \\\shline
    \end{tabular}
    \caption{\small{Quantitative comparison with baselines on TAP-Vid (left) and D3D-HOI (right).}}
    \label{table:baseline}
\end{table}

\begin{table}[t]
    \centering
    \tablestyle{6pt}{1.1}
    \scriptsize
    \setlength\tabcolsep{5pt}
    \begin{tabular}{l|cccc|c}
    \shline
    Ablations & AJ ($\uparrow$) & AD ($\downarrow$) & $< \delta^x_{avg}$ ($\uparrow$) \\
    \hline
    w/o visible region mask & 29.9 & 13.0 & 39.8 \\
    w cycle-consistency mask & 32.4 & 13.2 & 42.6 \\\arrayrulecolor{tablegray}\hline
    w/o feature-metric loss & 32.5 & \textbf{12.3} & 42.9 \\
    w/o photometric loss & 32.0 & 12.4 & 42.1 \\\arrayrulecolor{tablegray}\hline
    w/o flow regularizer & 24.7 & 18.8 & 34.3 \\
    w smoothness regularizer & 32.8 & 13.4 & 43.2 \\\arrayrulecolor{tablegray}\hline
    w/o semantic prior & 29.1 & 22.7 & 40.2 \\\arrayrulecolor{tablegray}\hline
    Full model & \textbf{33.2} & \textbf{12.3} & \textbf{43.5} \\ \arrayrulecolor{black}\shline
    \end{tabular}
    \captionof{table}{\small{Ablation study on model design choices.}}
    \vspace{-0.2in}
    \label{table:ablation}
\end{table}

\vspace{1mm}
\textbf{D3D-HOI} is a video dataset with annotations of 3D object pose and part motion during human-object interaction. 
We filter out the videos with severe occlusions and noisy annotations, resulting in a subset of 159 videos and 4 object categories, all with revolute joints. We choose two frames from each video that capture half of the articulated motion as inputs. We estimate pixel correspondence on RGB frames and get their 3D points using depth images. 
We use the least square algorithm~\cite{bjorck1996numerical} to estimate the articulation parameters using the predicted 3D point correspondence. We evaluate \textit{angle}, \textit{position} and \textit{state} errors of articulation parameters. To evaluate the predicted correspondence without ground truth annotation, we transform the points of the source image with the ground truth articulation parameters and compute the \textit{distance} between 
the transformed points and the corresponding points in the target image.

\begin{figure}
    \centering
    \includegraphics[width=0.98\linewidth]{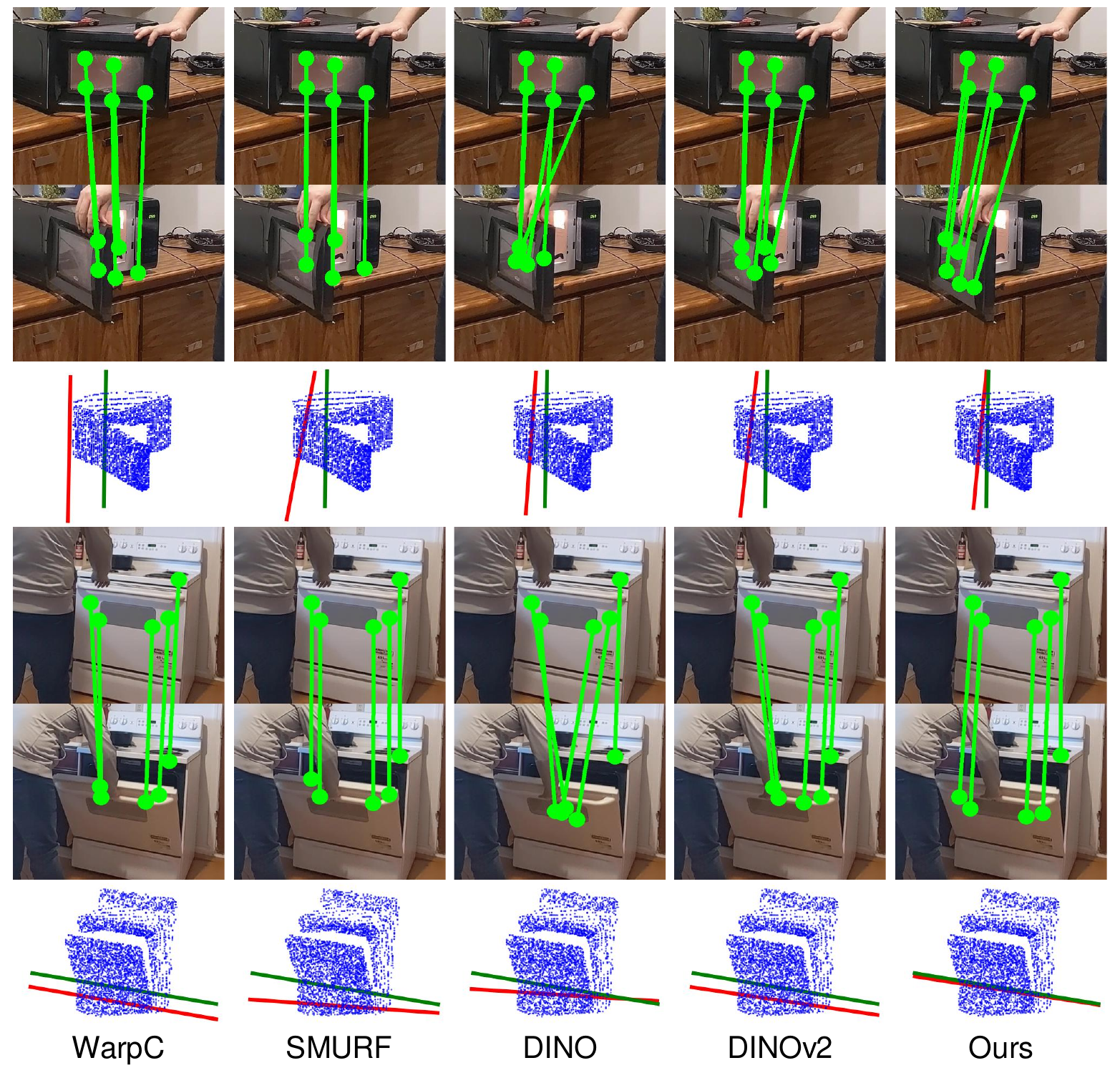}
    \vspace{-0.1in}
    \caption{\small{\textbf{Visualization on D3D-HOI dataset.} The top two rows show pixel-level correspondence. The last row shows articulation estimation results, where the green and red lines are ground truth and estimation.}}
    \vspace{-0.2in}
    \label{fig:articulation-qual}
\end{figure}

\vspace{1mm}
\noindent\textbf{Comparison with Baselines.} We evaluate the accuracy of predicted correspondence on TAP-Vid. As shown in Tab.~\ref{table:baseline} (left) and Fig.~\ref{fig:tapvid-qual}, our model \modelname{} outperforms baselines on all metrics with a significant margin. In detail, DINO and DINOv2 achieve a slightly worse AD than our model while getting much worse AJ and $<\delta^x_{avg}$. The reason is that they can only provide coarse semantic correspondence for relating points with similar semantics, leading to coarse correspondence. 
Meanwhile, WarpC, trained for accurate dense matching, achieves slightly better AJ than DINO with a much worse AD. 
As shown in Fig.~\ref{fig:tapvid-qual}, SMURF excels in matching pixels with little movement between the two frames and fails to establish long-range correspondence. Therefore, it has the best AJ and the worst AD among the baselines.

\subsection{Evaluation of Fine-grained Correspondence}

\noindent\textbf{Ablation Study.} To validate the design choices of \modelname{}, we conduct comprehensive ablation studies on the TAP-Vid dataset. The quantitative results are shown in Tab.~\ref{table:ablation}.

First, we evaluate the importance of the visible region discovery (Sec.~\ref{sec: visible_region_discovery}) by applying the losses on the entire image. We observe that applying losses on regions invisible in the target frame forces the network to find correspondence for pixels with no genuine matching. It provides a false supervisory signal, leading to worse performance. We also implement a cycle-consistency-based loss mask, which also gives inferior performance. Then, we evaluate the effect of feature-metric and photometric loss (Sec.~\ref{sec: loss}). Both ablated versions give a worse AJ, indicating both losses contribute to the accuracy of the estimated correspondence. In addition, we also investigate the flow regularizer (Sec.~\ref{sec: loss}). Removing the flow regularizer leads to a significant performance drop. Replacing the proposed distance consistency loss with a smoothness loss also leads to inferior performance, validating the efficacy of the distance consistency loss. Finally, we validate the importance of semantic prior by removing the matching candidate mask (Sec.~\ref{sec:flow}), which leads to a considerable drop in performance.

\subsection{Evaluation of Articulation Estimation}

As shown in Tab.~\ref{table:baseline} (right), our model gives better performance over the baselines on all metrics with significant margins. Visualizations in Fig.~\ref{fig:articulation-qual} show that WarpC cannot establish correct correspondences. SMURF tends to match points in the source frame to the points with the same coordinates in the target frame, while the DINO series can only match points with similar semantics. In contrast, \modelname{} demonstrates much more accurate correspondence prediction. The reason is that \modelname{} preserves the local structure of neighboring pixels during dense correspondence prediction, thanks to the self-supervised flow training with distance consistency regularizer. The accurate correspondence further improves the performance of articulation estimation.

\begin{table}[t]
    \centering
    \tablestyle{6pt}{1.1}
    \scriptsize
    \setlength\tabcolsep{5pt}
    \vspace{0.06in}
    \begin{tabular}{l|cccc|c}
    \shline
    Method & Monkey & Sloth & Peg Insertion \\
    \hline
    DON~\cite{florence2018dense} & 4/10 & 2/10 & 1/10 \\
    DINOv2~\cite{Oquab2023DINOv2LR} & 5/10 & 8/10 & 4/10 \\
    \modelname{} (ours) & \textbf{10}/10 & \textbf{9}/10 & \textbf{8}/10 \\ \shline
    \end{tabular}
    \captionof{table}{\small{Quantitative results of zero-shot goal-conditioned object manipulation.}}
    \vspace{-0.25in}
    \label{table:manipulation}
\end{table}

\subsection{Evaluation of Goal-Conditioned Object Manipulation}

We demonstrate that \modelname{} can be applied to zero-shot goal-conditioned object manipulation. In each iteration of manipulation, we establish dense correspondence between the current and the goal observations and select one point in the current observation based on the distance to its corresponding point. Then, we back-project the selected point and its corresponding target point into 3D space, producing a manipulation action to move the object closer to the goal.

\begin{figure}
    \centering
    \vspace{0.05in}
    \includegraphics[width=\linewidth]{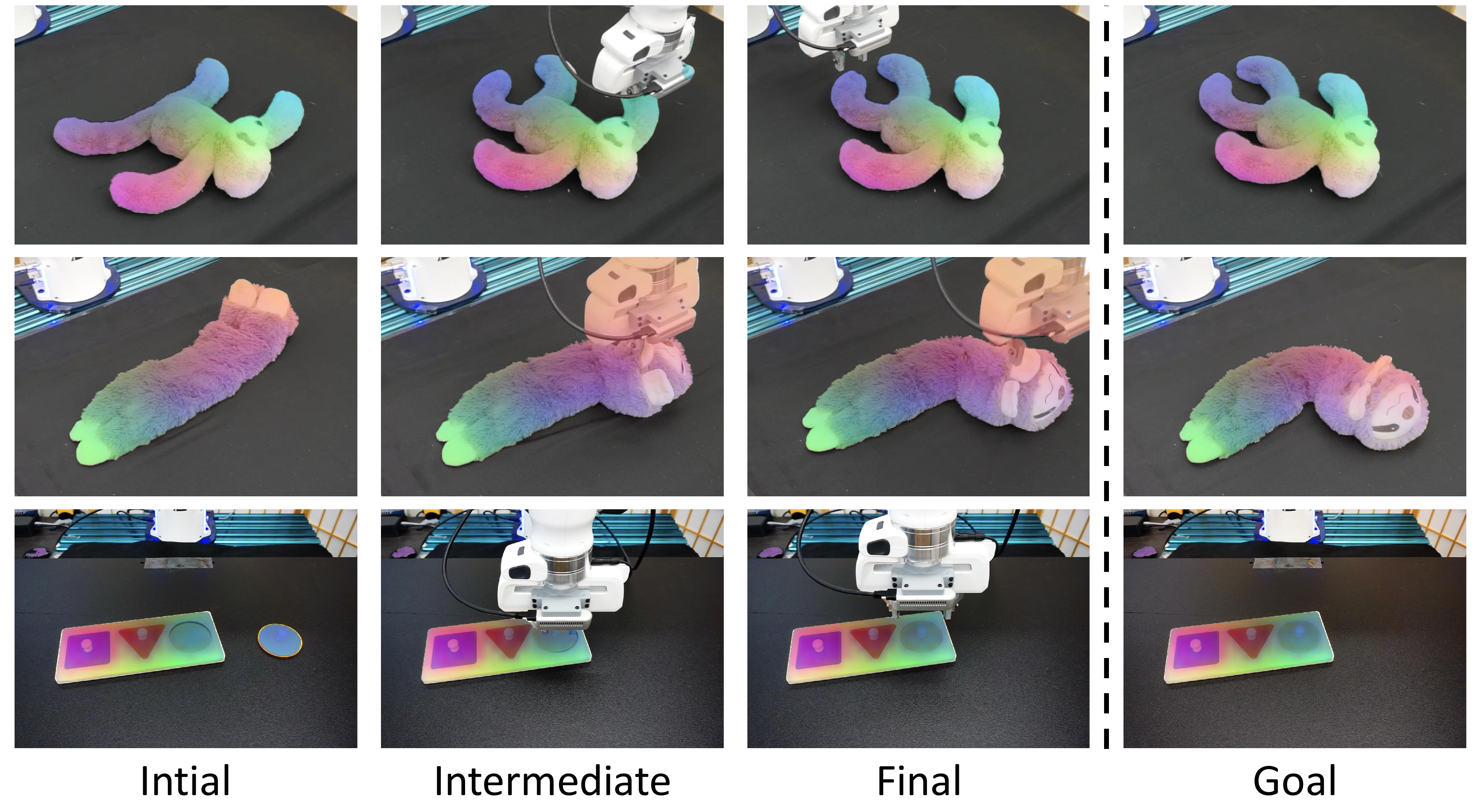}
    \vspace{-0.2in}
    \caption{\small{Illustrations of \modelname{} in goal-conditioned manipulation tasks: ``Monkey'', ``Sloth'', and ``Peg Insertion'', ordered from top to bottom. We visualize the dense correspondence of manipulated objects using PCA of image features.}}
    \vspace{-0.25in}
    \label{fig:manipulation}
\end{figure}

We conduct quantitative experiments on manipulating two deformable objects and one peg insertion task as in Fig.~\ref{fig:manipulation}. We compare with DINOv2~\cite{Oquab2023DINOv2LR}, the strongest baseline in the other two evaluations, and DenseObjectNets (DON)~\cite{florence2018dense}, a well-established dense visual descriptor for robotics manipulation. Since DON produces category-level descriptors and we want to test the zero-shot generalizability of the visual correspondence model, we use the official model weights of DON pretrained on caterpillar observations, which are closest to our tested deformable objects.

We evaluate the success rate of goal-conditioned manipulation driven by these correspondence models. For the two deformable objects, we follow ACID~\cite{shen2022acid} and use Chamfer Distance as the success metric. We use a 7-DoF Franka Emika Panda arm with an Intel RealSense D435i RGBD camera to execute the manipulation action. 

As shown in Tab.~\ref{table:manipulation}, DON gets relatively low success rates. It's because DON fails to generalize to novel objects, especially the blocks in the peg insertion task. DINOv2 achieves a decent success rate in deformable object manipulation but fails in peg insertion tasks where precise manipulation is required. Accurate visual correspondence from \modelname{} leads to fine-grained actions, giving the highest success rate.

\section{Conclusion}

In this work, we present \modelname{}, a self-supervised learning approach for dense visual correspondence. \modelname{} blends the advantages of self-supervised optical flow and semantic feature learning, establishing robust, dense correspondences between image pairs that capture scenes undergoing signification transformations. Results show that \modelname{} predicts more accurate point-level correspondence over baselines. Furthermore, we demonstrate the applicability of \modelname{} to robotic tasks such as articulation estimation and zero-shot goal-conditioned manipulation. These results manifest the potential of dense visual correspondence in robotics perception.

In the future, we would like to scale up the training of \modelname{} to more diverse in-the-wild images, including arbitrary pairs of images with any common content, leveraging the full potential of our self-supervised training paradigm.

\section*{ACKNOWLEDGMENT}
We would like to thank Yifeng Zhu for his help with real robot experiments and Alan Sullivan and Yue Zhao for helpful discussions. This work has been partially supported by NSF CNS-1955523, the MLL Research Award from the Machine Learning Laboratory at UT-Austin, and the Amazon Research Award.

\bibliographystyle{IEEEtran}
\bibliography{references}

\end{document}